\documentclass[journal]{IEEEtran}
\usepackage{amsmath,amsfonts}
\usepackage{algorithmic}
\usepackage{array}
\usepackage{textcomp}
\usepackage{stfloats}
\usepackage{url}
\usepackage{verbatim}
\usepackage{graphicx}

\usepackage{color}
\usepackage{subfigure} % conflicting with subfig
\usepackage{booktabs}
\usepackage{multirow}
\usepackage{bm}

\def\BibTeX{{\rm B\kern-.05em{\sc i\kern-.025em b}\kern-.08em
    T\kern-.1667em\lower.7ex\hbox{E}\kern-.125emX}}
\usepackage{balance}

\newcommand{\ie}{\textit{i}.\textit{e}., }
\newcommand{\eg}{\textit{e}.\textit{g}., }

\begin{document}
\title{Learning to Agree on Vision Attention for Visual Commonsense Reasoning}

\author{Zhenyang Li,
        Yangyang Guo,~\IEEEmembership{Member,~IEEE},
        Kejie Wang,
        Fan Liu,~\IEEEmembership{Member,~IEEE},
        Liqiang Nie,~\IEEEmembership{Senior Member,~IEEE},
        Mohan Kankanhalli,~\IEEEmembership{Fellow,~IEEE}
\IEEEcompsocitemizethanks{
% \IEEEcompsocthanksitem This work is supported by the National Natural Science Foundation of China, No.:U1936203. 
\IEEEcompsocthanksitem Zhenyang Li and Kejie Wang are with Shandong University, China. E-mail: \{zhenyanglidz, kjwang.henry\}@gmail.com.
\IEEEcompsocthanksitem Yangyang Guo, Fan Liu and Mohan Kankanhalli are with National University of Singapore, Singapore. E-mail: \{guoyang.eric, liufancs\}@gmail.com, mohan@comp.nus.edu.sg.
\IEEEcompsocthanksitem  Liqiang Nie is with Harbin Institute of Technology (Shenzhen), China. E-mail: nieliqiang@gmail.com.
}}

\markboth{IEEE TRANSACTIONS ON MULTIMEDIA}%
{Learning to Agree on Vision Attention for Visual Commonsense Reasoning}

\maketitle

\begin{abstract}
Visual Commonsense Reasoning (VCR) remains a significant yet challenging research problem in the realm of visual reasoning. 
A VCR model generally aims at answering a textual question regarding an image, followed by the rationale prediction for the preceding answering process. 
Though these two processes are sequential and intertwined, existing methods always consider them as two independent matching-based instances. 
They, therefore, ignore the pivotal relationship between the two processes, leading to sub-optimal model performance. 
This paper presents a novel visual attention alignment method to efficaciously handle these two processes in a unified framework. 
To achieve this, we first design a re-attention module for aggregating the vision attention map produced in each process. 
Thereafter, the resultant two sets of attention maps are carefully aligned to guide the two processes to make decisions based on the same image regions. 
We apply this method to both conventional attention and the recent Transformer models and carry out extensive experiments on the VCR benchmark dataset. 
The results demonstrate that with the attention alignment module, our method achieves a considerable improvement over the baseline methods, evidently revealing the feasibility of the coupling of the two processes as well as the effectiveness of the proposed method.
\end{abstract}


\section{Introduction}\label{sec:intro}
\begin{figure}[!t]
  \centering
  \includegraphics[width=1.0\linewidth]{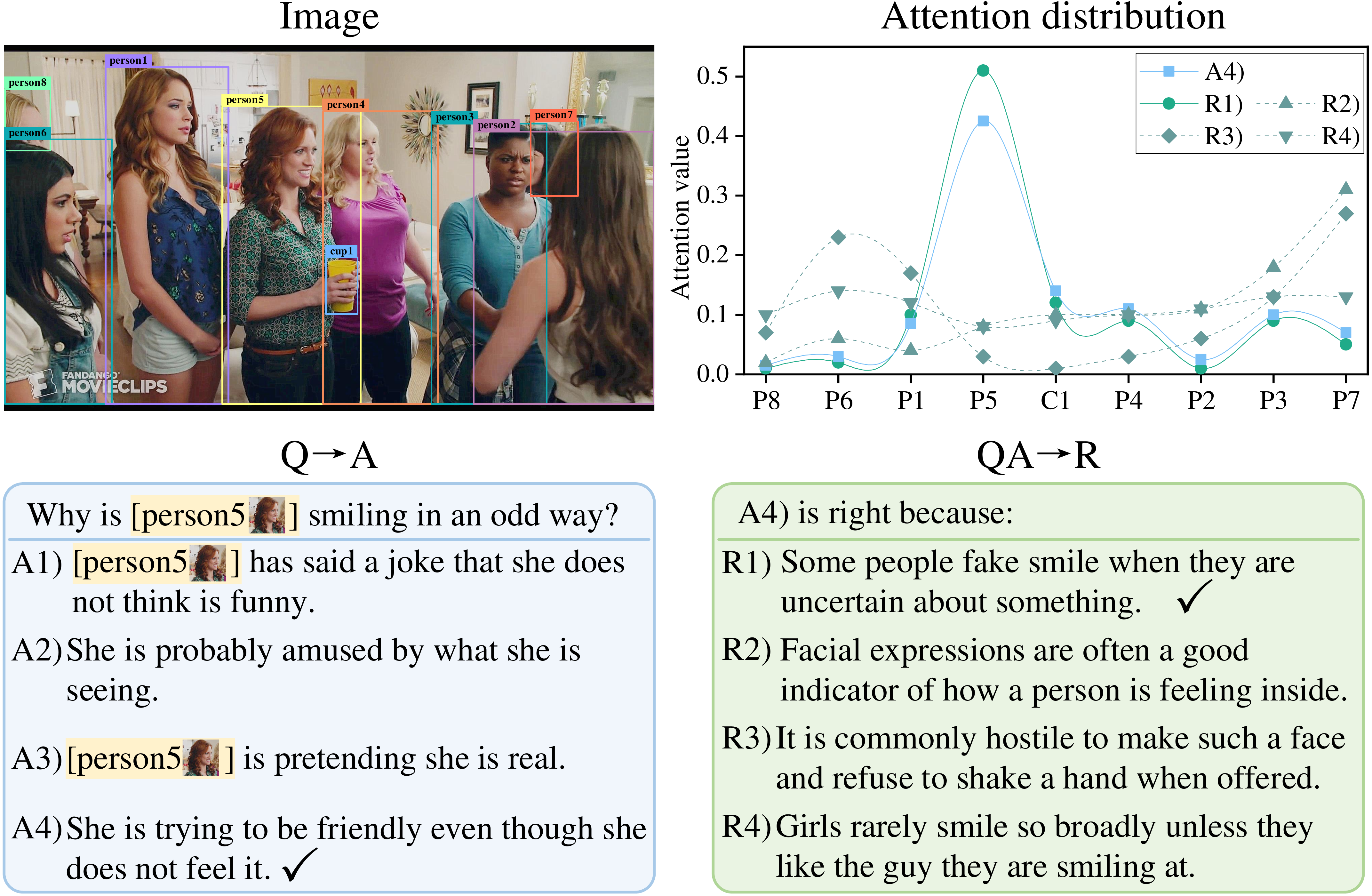}
  \caption{An instance of the VCR task and the attention distribution over image objects. 
  There are two sequential processes in VCR: Q$\rightarrow$A and QA$\rightarrow$R. 
  We propose to steer the attention from the right rationale (R1) to be similar to the ground-truth answer (A4) while making other negative pairs dissimilar (such as R2 and A4).}\label{fig:teaser}
\end{figure}

\IEEEPARstart{V}{isual} Question Answering (VQA) has received increasing interest over the past few years, which requires correctly answering natural language questions about a given image~\cite{VQA1}. 
Despite significant progress having been made, existing VQA benchmarks mainly focus on answering simple recognition questions (\eg \emph{how many} or \emph{what color}), while the explanation of the question answering is often ignored. 
To close this gap, Visual Commonsense Reasoning (VCR) has recently been presented as a challenge for researchers~\cite{R2C}.
Specifically, beyond answering the cognition-level questions (Q$\rightarrow$A) as canonical VQA does, VCR further prompts to provide a rationale for the correct answer (QA$\rightarrow$R) (see Figure~\ref{fig:teaser} for an example).

In fact, VCR poses more challenges than VQA, which can be seen in the following two aspects: 
1) on the data side -- the images in the VCR dataset describe more complex situations in the real world, and the questions are rather challenging and require high-level visual reasoning capabilities (e.g., \emph{why} or \emph{how}). 
And 2) on the task side -- it is hard to simultaneously figure out the right answer and its right rationale. 
Typically, VCR models first predict the answer, based on which the rationale can then be selected from the candidates.
As the question answering has proved to be non-trivial for traditional VQA models~\cite{VQA2, language-bias1, language-bias2}, finding the right rationale simultaneously is even more difficult.

Current VCR methods mostly enhance the visual understanding with the given query\footnote{Either question or question with the correct answer (see Figure~\ref{fig:teaser}).} (\ie to overcome the first challenge), and could be roughly divided into two categories: the first one is to leverage the intra-modality correlations to enhance separate feature learning, and inter-modality ones between vision and linguistics to correctly reason~\cite{HGL, TAB-VCR, CCN}; the other is to employ large external datasets to pre-train a general multi-modal model, and then transfer it to VCR for learning a better joint representation of the image and text~\cite{Unicoder, ViLBERT, UNITER}.
Although these methods have achieved promising results, they are all still limited by a common problem -- the Q$\rightarrow$A and QA$\rightarrow$R are handled independently, with the inherent relationship between these two processes being ignored. The second challenge thus remains unsettled, resulting in the sub-optimal performance of these methods.

In general, separately treating these two processes would inevitably lead VCR to degenerate into two VQA instances\footnote{Note that for QA$\rightarrow$R, the `question' to a VQA model now becomes the original question appended with the right answer. And the corresponding candidate `answer' set is accordingly composed of several rationales.}.
For each given question, existing methods consider Q$\rightarrow$A and QA$\rightarrow$R as two independent cases, rather than two sequential reasoning processes of one question.
This makes Q$\rightarrow$A and QA$\rightarrow$R disjoint, deviating from the original intention of VCR.
Moreover, answering and justification are actually consistent and coherent in human cognition: \emph{the human brain would capture and leverage the same evidence from the image for the two processes}.
Consequently, a better way to deal with VCR is to handle both via common cues. Thanks to the shared information provided by the same image, we align the visual attention of these two for better coordination. 

To seamlessly bridge the two processes, in this paper, we propose an a\textbf{G}ree on v\textbf{IS}ion a\textbf{T}tention model (dubbed GIST) for simple yet effective visual reasoning in VCR. 
Our method is composed of two consecutive modules. 
First, we design a novel re-attention module to model the fine-grained multi-modal interactions between queries and images. 
In particular, our attention module is two-stage: the first stage calculates the relevance between image regions and textual query tokens; the second stage combines the attention map of each token to obtain one single attention vector for each process.
Secondly, we align the produced attention maps over the given image, guiding the two processes to ‘look at the same image regions’.
Specifically, our method enforces an alignment loss between the attention maps from two processes, aiming to obtain similar attention maps during training.

To test the effectiveness of our method, we conduct extensive experiments on the VCR dataset. 
We test our GIST method on both the vanilla visual attention models as well as the most recent Vision Language Transformers (VL-Transformers)~\cite{VL-BERT, UNITER, villa}.
Both quantitative and qualitative results demonstrate that our method can significantly outperform the baseline method, confirming the utility of aligning vision attention of the two intertwined processes.

In summary, the contribution of this paper is threefold:
\begin{itemize}
\item This work addresses the question answering and rationale prediction in VCR with a unified framework. In particular, we propose an attention alignment module to guide Q$\rightarrow$A and QA$\rightarrow$R to make predictions with the same visual evidence.
\item We design a novel GIST method that skillfully aligns the visual attention from the two processes.
We apply this method to both the vanilla visual attention model and the recent VL-Transformers.
\item We conduct extensive experiments on the VCR dataset to demonstrate the effectiveness of our method. The code has been released\footnote{https://github.com/SDLZY/VCR\_Align.}. 
\end{itemize}
\section{Related Work}
\label{sec:related_work}
\subsection{Visual Commonsense Reasoning}
Conventional VQA mainly focuses on the recognition capability of models~\cite{VQA1, VQA5}. 
Until recently, VCR has emerged to study commonsense understanding, putting forward higher requirements in AI systems' cognitive reasoning abilities.
To address this task, many approaches have been proposed, which can be roughly divided into the following two categories.

The initial methods endeavor to devise task-specific architectures for VCR. 
Some of them explore a variety of sophisticated reasoning structures (\eg holistic attention mechanism) to construct interactions between the image and text.
For instance, R2C~\cite{R2C} performs three inference steps -- grounding, contextualization, and reasoning, to effectively solve the visual cognition problem. 
Inspired by the neuronal connectivity of the brain, CCN~\cite{CCN} designs a graph method to globally and dynamically integrate the local visual neuron connectivity;
HGL~\cite{HGL} integrates the intra-graph and inter-graph to bridge the vision and language modalities.
In addition, as some questions cannot be directly answered from only the image information, external commonsense knowledge is also exploited in the cross-modal reasoning process~\cite{KVL-BERT}. 

Another prevalent stream is to apply VL-Transformers to the downstream VCR for both Q$\rightarrow$A and QA$\rightarrow$R~\cite{VL-BERT, Unicoder}.
For example, UNITER~\cite{UNITER} designs four novel pre-training tasks with conditional masking to learn universal image-text representations for various downstream multi-modal tasks. 
ViLBERT~\cite{ViLBERT} applies a dual stream fusion encoder to process visual and textual inputs in separate streams, followed by the modality interactions with co-attentional Transformer layers. 
MERLOT RESERVE~\cite{MERLOT_RESERVE} first learns task-agnostic representations through sound, language, and vision of videos~\cite{videos}, and then transfers these features to the downstream VCR task. 

Although both categories of methods have achieved improved results, they all view the two processes in VCR as two independent VQA instances. 
As a result, the critical correlations between these two are ignored, resulting in weak visual reasoning. 
In this work, we propose an effective attention alignment framework to bridge these two processes directly.

\subsection{Attention in Visual Question Answering}
The past few years have witnessed increasing growth in the research area of VQA. 
Among the existing methods, visual attention-based ones have demonstrated substantial advantages in feature learning. 
Traditional approaches focus more on the correlation learning between each image region and the whole question sentence~\cite{VQA5, VQA6}.
Thereafter, the co-attention mechanism jointly performs the question-guided attention over image regions and the image-guided attention over question words~\cite{Co-Attention}.
In addition to these top-down attention methods, BUTD~\cite{BUTD1, BUTD2} combines both the top-down and bottom-up attention, which first detects the salient objects inside an image and then leverages the top-down attention technique to locate the most relevant regions according to the given question.
VQA-HAT~\cite{Human-LikeAttention1} and Attn-MFH~\cite{Human-LikeAttention2} point out that the attention maps produced by VQA models are inconsistent with human cognition. They, therefore, attempt to design more explicitly visual reasoning methods and have achieved certain improvements.

\subsection{Vision-Language Transformers} \label{sec:pre-train-objective}
Transformers have achieved great success in the field of Natural Language Processing (NLP)~\cite{BERT} and Computer Vision (CV)~\cite{ViT}. Because of their effectiveness, transformers also attract much attention in multi-modal studies.
Based on how the vision and language branches are fused, current VL-Transformers can be roughly categorized into single-stream  (\eg UNIMO~\cite{UNIMO} and SOHO~\cite{SOHO}) and dual-stream cross-modal Transformers (\eg LXMERT~\cite{LXMERT} and ALBEF~\cite{ALBEF}). 
In general, VL-Transformers adopt a \emph{pretrain-then-finetune} learning paradigm: these models are firstly pre-trained on large-scale multi-modal datasets (such as Conceptual Captions~\cite{Conceptual_Captions}) for learning universal cross-modal representations, and then fine-tuned on downstream tasks by transferring their rich representations from pre-training. 
In particular, the pretext tasks play an important role in pre-training, where masked language modeling, masked region prediction, and image-text matching are extensively studied. 
The fine-tuning step mirrors that of the BERT model~\cite{BERT}, which includes a downstream task-specific input, output, and objective.
VL Transformers mainly help the following three groups of downstream tasks: cross-modal matching, cross-modal reasoning, and vision language generation~\cite{Survey_of_VLTransformer}. 
The first group focuses on learning cross-modal correspondences between vision and language, such as image text retrieval and visual referring expression. 
Reasoning ones require VL Transformers to perform language reasoning based on visual scenes, such as VQA. 
The last group aims to generate the targets of one modality given the other as input~\cite{X-LXMERT}. The desired visual or textual tokens are decoded in an auto-regressive generation manner.
\section{Proposed Method}
Based on task intuition and human cognition, the answering and reasoning processes in VCR should be made cohesive and consistent. 
Nevertheless, existing methods often treat them separately, rendering the commonsense reasoning less convincing. 
To tackle these two processes jointly, we resort to aligning the visual attention of question answering and rationale inference processes for a better collaborative connection and design our GIST method.

In the following, we first introduce the method's intuition and background knowledge, followed by our proposed visual attention alignment method. 
We then detail its implementation on the model with vanilla attention and the recent VL-Transformer with self-attention, respectively. 

\begin{figure}[!t]
  \centering
  \includegraphics[width=0.99\linewidth]{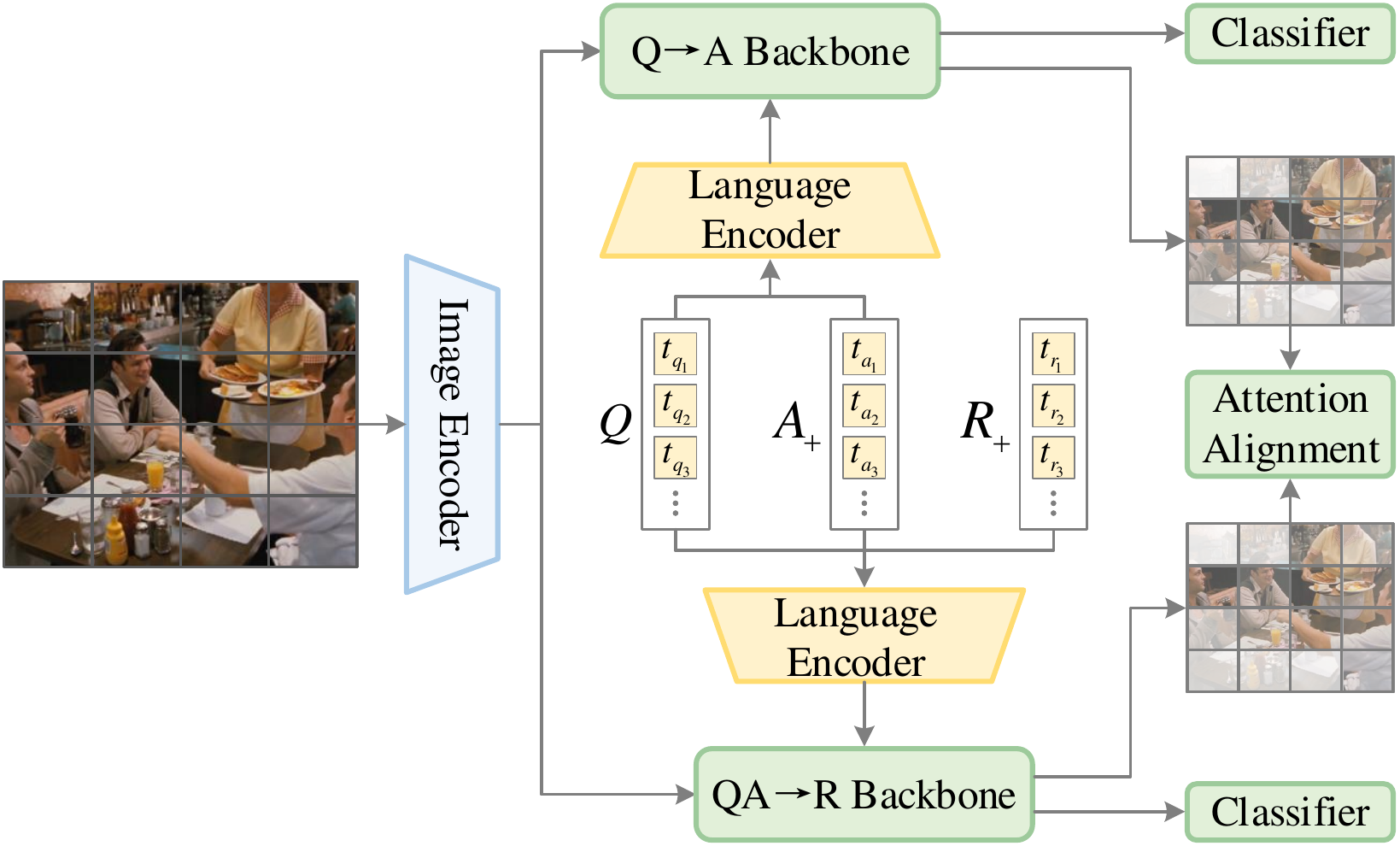}
  \caption{Pipeline of the proposed method. 
  Both Q$\rightarrow$A and QA$\rightarrow$R share the same image, which is encoded by the common image encoder.
  $t$ denotes the input tokens.
  The Q$\rightarrow$A takes the image, question, and answer as input, and produces its attention map.
  Similarly, we can also obtain the attention map from QA$\rightarrow$R.
  Our goal is to make these two attention maps as similar as possible, so that these two processes \emph{see} the same image regions for reasoning.} \label{fig:overview}
\end{figure}

\begin{figure*}[!t]
  \centering
  \includegraphics[width=1.0\linewidth]{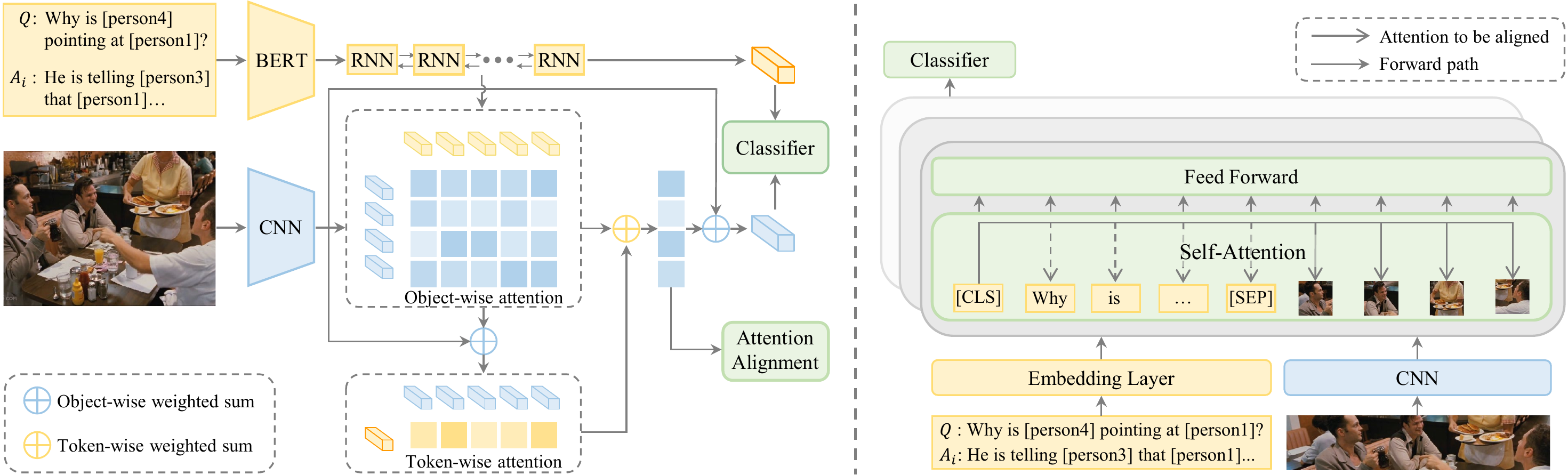}
  \caption{Architecture of our proposed method with vanilla attention network (Left) and VL-Transformers (Right). 
  For the vanilla attention model, we sequentially perform object-wise and token-wise attention to obtain the final attention maps for all the detected objects.
  Pertaining to the VL-Transformers, the attention weights are extracted from the self-attention operation.}\label{fig:vanilla_attention}
\end{figure*}

\subsection{Method Intuition}
Given an image $I$ and a question $Q$ about this image, the goal of visual commonsense reasoning is to both predict the right answer $A_{+}$, as well as the correct rationale $R_{+}$ (the right and false ones are denoted as $+$ and $-$, respectively). 
In general, VCR comprises two multiple-choice processes: question answering (Q$\rightarrow$A) and answer justification (QA$\rightarrow$R). 

\textbf{Q$\rightarrow$A} aims to predict the correct answer $A_{+}$ from a set of answer choices $\mathcal{A}$ to the given question $Q$ upon the image $I$, which can be achieved by,
\begin{equation}
\label{eq:objective_q2a}
    A_{+} = \mathop{\arg\max}\limits_{A_i\in{\mathcal{A}}} {f(I, Q \mid A_i)},
\end{equation}
where $f$ denotes the Q$\rightarrow$A model.
Specifically, both question $Q$ and candidate answer $A_i$ are expressed in terms of textual sentences, and the image $I$ consists of $n$ objects $\mathcal{O} = \{o_i\}_{i=1}^{N}$ detected by a Mask-RCNN model~\cite{MASK_RCNN}. 
Previous methods train $f$ by minimizing the cross entropy loss\footnote{We employ the single instance loss rather than the batch-wise one for simplicity.},
\begin{equation}
    \mathcal{L}_{Q \mapsto A} = - y[A_+] \log\frac{\exp{f(I, Q \mid A_+)}}{\sum\nolimits_{i=1}^{\mid \mathcal{A} \mid}\exp{f(I, Q \mid A_i)}},
\end{equation}
where $y[A_+]$ denotes the index of the ground-truth answer. 

\textbf{QA$\rightarrow$R} is different from Q$\rightarrow$A wherein the input is now composed of question $Q$ and its correct answer $A_{+}$ (a straightforward way is to concatenate these two together). 
The objective of QA$\rightarrow$R is to select the correct rationale $R_{+}$ from a rationale set $\mathcal{R}$ (see Figure~\ref{fig:teaser}),
\begin{equation}
\label{eq:objective_qa2r}
    R_{+} = \mathop{\arg\max}\limits_{R_i\in{\mathcal{R}}} {g(I, Q, A_{+} \mid R_i)},
\end{equation}
where $g$ denotes the QA$\rightarrow$R model.
The optimization function is similarly defined as follows,
\begin{equation}
    \mathcal{L}_{{QA \mapsto R}} = - y[R_+] \log\frac{\exp{g(I, Q, A_{+} \mid R_+)}}{\sum\nolimits_{i=1}^{\mid \mathcal{R} \mid}\exp{g(I, Q, A_{+} \mid R_i)}},
\end{equation}
where $y[R_+]$ represents the ground-truth rationale for the given image and question. 

Note that both $f$ and $g$ share identical structures and are often trained separately~\cite{CCN, VCR_TMM1}. 
To bridge these two connected processes, we propose a visual attention alignment module in the next.

\subsection{Visual Attention Alignment} \label{sec:align}
Inspired by human cognition, we argue that the visual evidence exploited in answering and justification should be the same. 
As these two processes share the same image (see Equation~\ref{eq:objective_q2a} and Equation~\ref{eq:objective_qa2r}), the fundamental visual attention thus offers a natural bridge for connecting them. 
In view of this, we propose to align the visual information based on the attention maps calculated by $f$ and $g$. 

Visual attention is an integral component for current vision-language models~\cite{BUTD1, VCR_TMM2}. 
A typical VCR model often involves a visual attention module on the object set $\mathcal{O}$. 
In particular, the visual attention learns a set of attention score $\{c_i\}_{i=1}^N$ for each image object, where $\sum_{i=1}^N c_i = 1$.
A large $c_i$ usually represents that the $i$-th object has more influence for answering the current question.
As discussed earlier, there are two sets of attention maps: $\mathcal{C}_{Q \mapsto A}$ -- the attention weights from model $f$, and $\mathcal{C}_{QA \mapsto R}$ -- the attention weights from model $g$.
To achieve the goal that $f$ and $g$ employ the similar image regions, our idea is to learn another module that drives $\mathcal{C}_{Q \mapsto A}$ and $\mathcal{C}_{QA \mapsto R}$ closer.
Specifically, we implement this by pulling the similarity of attention maps from the correct answer ${A_+}$ and the correct rationale ${R_+}$, while pushing away other negative pairs. 

In this paper, we propose an attention alignment loss to make the Q$\rightarrow$A and QA$\rightarrow$R learn to agree on visual regions. 
We formalize our attention alignment loss below,
\begin{equation}
    \begin{cases}
        \mathcal{L}_{Q \mapsto A}^{Att} = -y[A_+] \log\frac{\exp{s(\mathcal{C}_{Q \mapsto A}^{A_+}, \mathcal{C}_{QA \mapsto R}^{R_+})}}{\sum\nolimits_{i=1}^{\mid \mathcal{A} \mid}\exp{s(\mathcal{C}_{Q \mapsto A}^{A_i}, \mathcal{C}_{QA \mapsto R}^{R_+})}},    \\
        \mathcal{L}_{QA \mapsto R}^{Att} = -y[R_+] \log\frac{\exp{s(\mathcal{C}_{QA \mapsto R}^{R_+}, \mathcal{C}_{Q \mapsto A}^{A_+})}}{\sum\nolimits_{i=1}^{\mid \mathcal{R} \mid}\exp{s(\mathcal{C}_{QA \mapsto R}^{R_i}, \mathcal{C}_{Q \mapsto A}^{A_+})}},    \\
        \mathcal{L}_{Align} = \mathcal{L}_{Q \mapsto A}^{Att} + \mathcal{L}_{QA \mapsto R}^{Att} ,
    \end{cases}
\end{equation}
where $\mathcal{C}_{Q \mapsto A}^{A_i}$ represents the attention weights from model $f$ according to the $i$-th answer $A_i$; $\mathcal{C}_{QA \mapsto R}^{R_i}$ denotes the attention weights from model $g$ according to the $i$-th rationale $R_i$, and $s(\cdot, \cdot)$ is a similarity function.
We detail how we implement $s(\cdot, \cdot)$ in the following:
\begin{itemize}
    \item One intuitive approach to measure the similarity between two sets of attention weights is the dot product. 
    We refer to this method as \emph{Align-Dot},
    \begin{equation*}
        s(\mathbf{c}_p, \mathbf{c}_t) = \mathbf{c}_p^T \mathbf{c}_t.
    \end{equation*}
    \item Inspired by the \emph{learning to rank} models in the field of information retrieval~\cite{L2R}, we use the list-wise approach to align the ranking of attention weights and name this method \emph{Align-Rank}.
    To this end, we first obtain the permutations of the two attention vectors prediction $\boldsymbol{\pi}_p$ and target  $\boldsymbol{\pi}_t$. 
    Thereafter, we employ the NDCG metric optimization~\cite{NDCG, App_Rank} in our experiments:
    \begin{equation}
    \begin{aligned}
        s(\boldsymbol{\pi}_p, \boldsymbol{\pi}_t)   &= NDCG(\boldsymbol{\pi}_p, \boldsymbol{\pi}_t) \\
                                                    &= Z_{N}^{-1}\sum_{i=1}^{N}\frac{G(\pi_{pi})}{\log(1+\pi_{ti})},
    \end{aligned}
    \end{equation}
    where $G(\pi_{pi})$ denotes a gain function, \eg $G(\pi_{pi}) = 2^{\pi_{pi}} - 1$, $Z_{N}^{-1}$ acts as the maximum of $G(\pi_{pi}) / \log(1+\pi_{ti})$, \ie the value when the predicted attention permutation of $\boldsymbol{\pi}_p$ is the same as the target one $\boldsymbol{\pi}_t$.
    Nevertheless, directly optimizing NDCG with back-propagation is impossible due to its non-differentiable nature.
    To approach this problem, we then smooth the permutation function with~\cite{App_Rank},
    \begin{equation}
        \hat{\pi}_{pi} = 1 + \sum_{j\neq i}\frac{\exp\{-\alpha(c_{pi} - c_{pj})\}}{1 + \exp\{-\alpha(c_{pi} - c_{pj})\}},
    \end{equation}
    where $\alpha$ is a hyper-parameter, and $c_{pi}$ denotes the attention value of the $i$-th object.
\end{itemize}

By combining the aforementioned loss functions, the final objective becomes,
\begin{equation}
    \mathcal{L} = \mathcal{L}_{{Q \mapsto A}} + \mathcal{L}_{{QA \mapsto R}} + \lambda \mathcal{L}_{Align}
\end{equation}
where $\lambda$ is a trade-off hyper-parameter.

Our method is applicable to both vanilla visual attention models as well as the most recent VL-Transformers. 
In the following, we will show its implementation on these two typical models.

\subsection{Application on the Vanilla Attention Model}
Before the prevalence of VL-Transformers in VCR, conventional methods all adopt the vanilla attention mechanism to focus on the most salient image regions based on the textual information.
In view of this, we intend to explore whether our proposed visual attention alignment method works under such settings.
Specifically, we take the TAB-VCR~\cite{TAB-VCR} as a typical baseline to implement our method.
Note that the two processes, \ie Q$\rightarrow$A and QA$\rightarrow$R share the same structure, we, therefore, use the Q$\rightarrow$A as an example as the QA$\rightarrow$R can be easily extrapolated.

\subsubsection{Overall Framework} In this subsection, we first introduce the overall framework of TAB-VCR. 

\textbf{Image \& Language Encoder.} Images in the VCR dataset~\cite{R2C} consist of objects detected by Mask-RCNN~\cite{MASK_RCNN}. 
We leverage this fine-grained information and utilize the pre-trained Convolutional Neural Network (CNN) model~\cite{ResNet} to extract the object features $\mathcal{O}=\{\boldsymbol{o}_1, ..., \boldsymbol{o}_N\}$. 

Pertaining to the textual input, we first concatenate the question $q$ and each answer $a_i$ and then employ a pre-trained word embedding to obtain the embeddings.
Note that each sentence in VCR includes both text and some object tags, as shown in Figure~\ref{fig:vanilla_attention}.
Following previous studies~\cite{R2C, TAB-VCR}, we take the word embeddings and the object features in their right order as input to a bidirectional RNN~\cite{GRU}.
Specifically, if an input token is a tag referring to an object $o_{t}$, and the corresponding object feature will be $\boldsymbol{o}_{t}$; otherwise, it is the embedding of the entire image.
The query and response can be encoded into a sequence of hidden states $\{\boldsymbol{h}_1, ..., \boldsymbol{h}_{M}\}$ by an RNN model:
\begin{equation} \label{eq:rnn}
    \boldsymbol{h}_t = \mathsf{RNN}([\boldsymbol{t}_i, \boldsymbol{o}_{t_i}]; \boldsymbol{h}_{t-1}),
\end{equation}
where $\boldsymbol{o}_{t_i}$ denotes the $t_i$-th object's feature if $\mathbf{t}_i$ is a tag otherwise the averaged feature of all objects.

\textbf{Classifier.} After the visual reasoning between the given question and image (which will be detailed in the next subsection), we treat VCR as a multi-class classification problem. 
In particular, we use the last hidden state $\boldsymbol{h}_M$ from RNN and the refined image feature $\hat{\boldsymbol{o}}$ as the representation of the question-answer pair and image, respectively.
Our classifier is implemented with a multi-layer perceptron (MLP)~\cite{MLP} to compute a score for the candidate answers:
\begin{equation} \label{eq:classifier}
    \hat{y_i} = \boldsymbol{W}_1\sigma({\boldsymbol{W}_0[\boldsymbol{h}_M, \hat{\boldsymbol{o}}]}),
\end{equation}
where $\sigma$ is the LeakyReLU~\cite{ResNet} activation function, and $\boldsymbol{W}_0$ and $\boldsymbol{W}_1$ are the learned weight matrices in the MLP.
We omit the bias vectors for simplicity. 

\subsubsection{Re-Attention} \label{sec:vanilla-re-att}
A typical model often extracts the visual features with a CNN model~\cite{cnn-mm} and takes the output $\mathcal{O}$ as the vision features for Equation~\ref{eq:rnn}.
Different from it, we propose a re-attention module to distribute distinctive attention weights, which serves as an important part of our attention alignment goal.
Specifically, our re-attention module consists of attention computation from two directions: object-wise and token-wise.
The former takes each token feature as a query and learns the attention weights for all the objects.
In contrast, token-wise attention collects the most informative signals of each token based on the fused token information from the former step. 

\textbf{Object-wise Attention.}
It is intuitive that for each token, different objects often contribute distinctively to the learning of the textual features.
For example, in Figure~\ref{fig:teaser}, the \emph{[person5]} object is more important than other objects for learning the token \emph{smiling} in the question.
In light of this, we employ the token to attend to each specific object as follows,
\begin{equation}
\begin{cases}
    \bar{\mathbf{c}}_{o_{t_i}} = \mathsf{MLP} (\mathbf{t}_i)^T \mathsf{MLP} ([\mathbf{o}_1, \mathbf{o}_2 \cdots \mathbf{o}_N]), \\
    \mathbf{c}_{o_{t_i}} = \mathsf{SoftMax} (\bar{\mathbf{c}}_{o_{t_i}}),\\
\end{cases}
\end{equation}
where $\mathbf{c}_{o_{t_i}}$ and $\bar{\mathbf{c}}_{o_{t_i}} \in \mathbb{R}^M$.
In this way, we can have $M$ attention maps over all the objects, where $M$ is the number of tokens.

\textbf{Token-wise Attention.}
After the object-wise attention operation, we then employ another attention module to estimate the importance of each token with respect to the overall textual feature. 
To implement this, we take the output from RNN model $\mathbf{h}_M$ as a query, and perform attention over all the token features,
\begin{equation} \label{eq:re-att1}
\begin{cases}
    \bar{\mathbf{c}}_t = \mathsf{MLP} (\mathbf{h}_M)^T \mathsf{MLP} ([\mathbf{h}_1, \mathbf{h}_2, \cdots, \mathbf{h}_M]), \\
    \mathbf{c}_t = \mathsf{SoftMax} (\bar{\mathbf{c}_t}),\\
\end{cases}
\end{equation}
where $\mathbf{c}_t$ and $\bar{\mathbf{c}_t} \in \mathbb{R}^M$.
Thereafter, we employ the attention weights, \ie the importance of tokens $\mathbf{c}_t$, to multiply that of the set of attention maps, to obtain the overall attention weights over objects $\hat{\mathbf{o}}$,
\begin{equation} \label{equ:final_att}
  \begin{aligned}
    \mathbf{c}_o    &= \sum_{i=1}^M c_{t_i} \times \mathbf{c}_{o_{t_i}}, \\
    \hat{\mathbf{o}}&= \sum_{i=1}^N c_{o_i} \times \mathbf{o}_i.
  \end{aligned}
\end{equation}

\subsubsection{Attention Alignment}
We then obtain the refined image feature in Equation~\ref{eq:classifier} and perform the final classifier for predicting the right answer or rationale.
Thereafter, following Equation~\ref{equ:final_att}, the two attention sets can be easily collected from $\mathbf{c}_o$.
One is for $\mathcal{C}_{Q \mapsto A}$, and the other is $\mathcal{C}_{Q \mapsto A}$.
The visual attention alignment from Sec.~\ref{sec:align} can thereby be performed.

\begin{table*}[htb]
    \centering
    \caption{Performance comparison on the validation and testing sets.  
    The best results are highlighted in bold.}
    \begin{tabular}{{r}|ccc|*{6}{c}}
    \toprule
    \multirow{2}{*}{Model} & \multirow{2}{*}{VQA} & \multirow{2}{*}{VCR} & \multirow{2}{*}{Transformer} & \multicolumn{2}{c}{Q$\rightarrow$A} & \multicolumn{2}{c}{QA$\rightarrow$R} & \multicolumn{2}{c}{Q$\rightarrow$AR} \\
                                                                            \cmidrule(lr){5-6}     \cmidrule(lr){7-8}     \cmidrule(lr){9-10}
                                     &            &            &            & valid  & test        & valid  & test        & valid  & test  \\
    \midrule
    Chance                           &            &            &            & 25.0   & 25.0        & 25.0   & 25.0        & 6.2    & 6.2   \\
    \midrule
    RevisitedVQA~\cite{RevisitedVQA} & \checkmark &            &            & 39.4   & 40.5        & 34.0   & 33.7        & 13.5   & 13.8  \\       
    BUTD~\cite{BUTD1}                & \checkmark &            &            & 42.8   & 44.1        & 25.1   & 25.1        & 10.7   & 11.0  \\
    MLB~\cite{MLB}                   & \checkmark &            &            & 45.5   & 46.2        & 36.1   & 36.8        & 17.0   & 17.2  \\
    MUTAN~\cite{MUTAN}               & \checkmark &            &            & 44.4   & 45.5        & 32.0   & 32.2        & 14.6   & 14.6  \\
    R2C~\cite{R2C}                   &            & \checkmark &            & 63.8   & 65.1        & 67.2   & 67.3        & 43.1   & 44.0  \\
    CCN~\cite{CCN}                   &            & \checkmark &            & 67.4   & 68.5        & 70.6   & 70.5        & 47.7   & 48.4  \\
    HGL~\cite{HGL}                   &            & \checkmark &            & 69.4   & 70.1        & 70.6   & 70.8        & 49.1   & 49.8  \\
    TAB-VCR~\cite{TAB-VCR}           &            & \checkmark &            & 69.5   & 70.5        & 71.6   & 71.6        & 50.1   & 50.8  \\
    VL-BERT~\cite{VL-BERT}           &            & \checkmark & \checkmark & 72.6   & 73.4        & 74.0   & 74.5        & 54.0   & 54.8  \\   
    UNITER~\cite{UNITER}             &            & \checkmark & \checkmark & 74.4   & 75.5        & 76.9   & 77.3        & 57.5   & 58.6  \\
    \midrule
    $\text{GIST}_{\text{Vanilla}}$   &            & \checkmark &            & \bf{70.5}& \bf{71.2} & \bf{72.5}& \bf{72.0} & \bf{51.5}& \bf{51.4} \\
    $\text{GIST}_{\text{VL-Transformer}}$&        & \checkmark & \checkmark & \bf{74.9}& \bf{75.6} & \bf{77.0}& \bf{77.5} & \bf{58.1}& \bf{58.8} \\
    \midrule
    Human                            &            & \checkmark &            & -      & 91.0        &-       & 93.0        &-       & 85.0        \\
    \bottomrule
    \end{tabular}
    \label{tab:overall}
\end{table*}

\subsection{Application on the VL-Transformer}
VL-Transformers have been widely studied over the past few years~\cite{VL_Transformers1, VCR_TMM3}. 
In this section, we show how our attention alignment method works under such self-attention settings.

\subsubsection{Overall Framework}
The structure of a typical single-stream VL-Transformer is illustrated in Figure~\ref{fig:vanilla_attention}.

\textbf{Transformer Encoder.}
As can be observed, the input sequence starts with a special classification token (\ie [CLS]), and then goes on with query and response elements, visual tokens, and ends with a special ending token (\ie [END]). 
A special separation token (\ie [SEP]) is introduced between every two parts of the input. 
Following the practice in BERT~\cite{BERT}, the input textual sentence is first split into tokens by the WordPiece tokenizer~\cite{WordPiece}, which are then transformed into vectors by the embedding layer.
Pertaining to the vision inputs, each token is represented as the detected object features, the same as that in the vanilla attention model.
After obtaining these embeddings, we then add them with the segmentation and position embeddings, so that the sequential information can be encoded into the transformer model.
Thereafter, these embeddings are inputted to several Transformer blocks, wherein each of them consists of a self-attention layer, a feedforward network, and some layer normalization operations. 

\textbf{Classifier.}
The features from both vision and language are fused and interacted with the Transformer model.
And it is expected that visual reasoning is also performed. 
At last, the final block's output of the [CLS] token is fed to a Softmax classifier to predict whether the given response is the correct choice.

\subsubsection{Re-Attention}
The key to a Transformer model is multi-head self-attention. 
Given query and key matrices, $\mathbf{Q}$ and $\mathbf{K}$, the single head attention is formally defined as follows,
\begin{equation} \label{equ:sa}
    \mathsf{SA}(\mathbf{Q}, \mathbf{K}) = \mathsf{SoftMax} (\frac{\mathbf{Q}^T\mathbf{K}}{\sqrt{d_k}}),
\end{equation}
where $d_k$ is the dimension of keys and values and acts as a scaling factor.
The more commonly used multi-head self-attention is formulated,
\begin{equation}
    \mathsf{MSA}(\mathbf{Q}, \mathbf{K}) = \mathsf{Concat} (h_1, \cdots, h_k) \mathbf{W},
\end{equation}
where $k$ is the number of attention heads, which is calculated by the $\mathsf{SA}$ function in Equation~\ref{equ:sa}.

One advantage of VL Transformers is that they offer holistic attention estimation with the [CLS] token.
In this way, we do not have to aggregate all the information from textual tokens in the first step, as Sec.~\ref{sec:vanilla-re-att} does.

\subsubsection{Attention Alignment}
To achieve the attention alignment goal, we first average all the information from the $k$ heads.
Thus, we obtain an $L$ layers attention map for the [CLS] token.
We then extract the attention to the visual tokens and have an attention matrix $\mathbf{C}_o \in \mathbb{R}^{L \times N}$.
To this end, we perform the visual attention alignment for each layer following the guidance of Sec.~\ref{sec:align}.

\section{Experiments}
\begin{figure}[htb]
  \centering
  \includegraphics[width=0.99\linewidth]{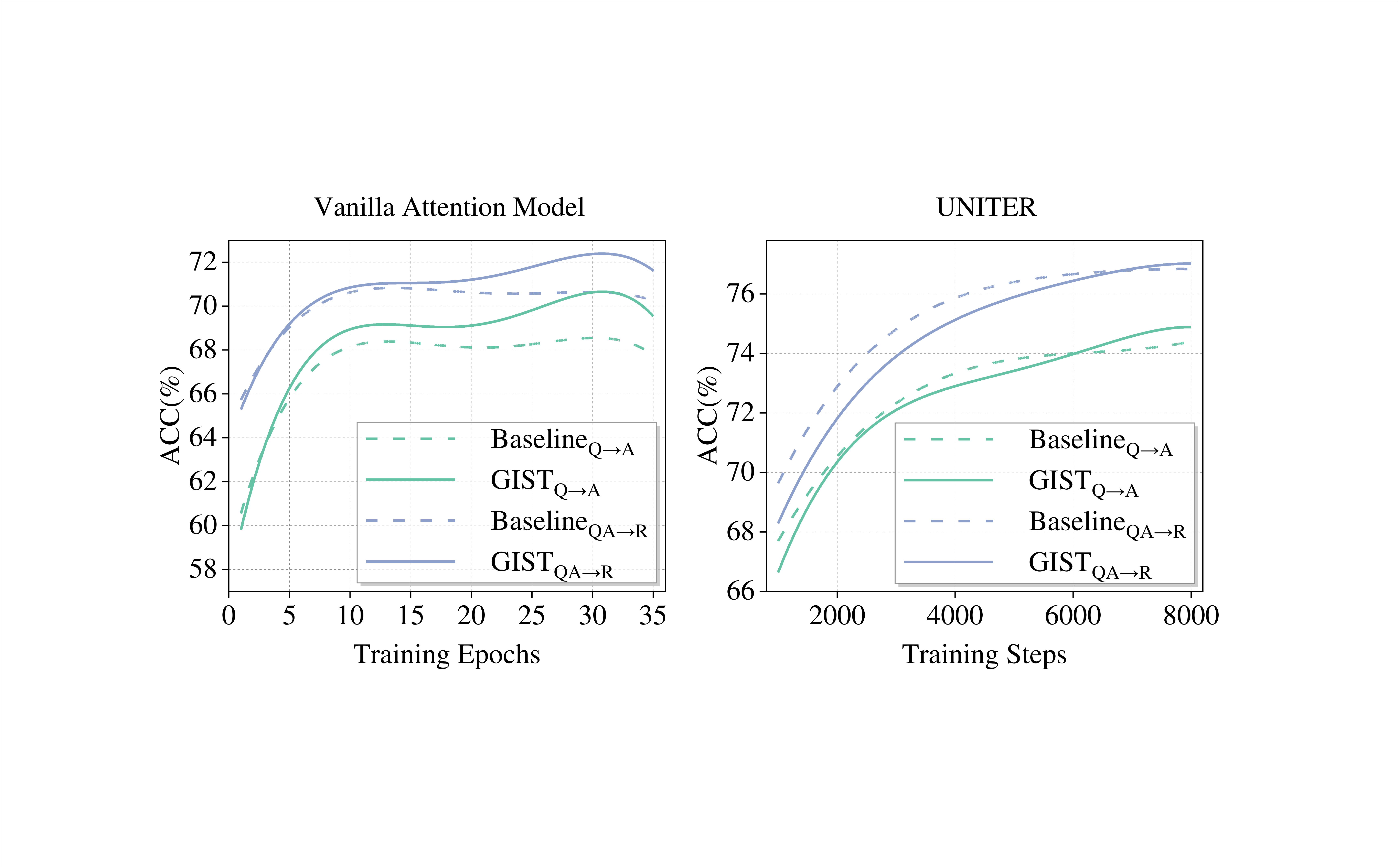}
  \caption{Convergence analysis of baselines and our GIST model.} \label{fig:convergence_analysis}
\end{figure}

\subsection{Datasets and Evaluation Protocols }
We conducted extensive experiments on the VCR benchmark, a large-scale dataset alongside this task.
The images are extracted from the movie clips in LSMDC~\cite{LSMDC} and MovieClips\footnote{youtube.com/user/movieclips.}, wherein the objects inside images are detected via the Mask-RCNN model~\cite{MASK_RCNN}.
We used the official dataset split, where the number of questions for training, validation, and testing are 212,923, 26,534, and 25,263, respectively. 
For each question, four answers are given with only one being correct, and there are also four rationale choices among which only one makes sense.

Regarding the evaluation metric, we used the popular classification accuracy  for Q$\rightarrow$A, QA$\rightarrow$R, and Q$\rightarrow$AR\footnote{For Q$\rightarrow$AR, the prediction is right only when both the answer and rationale are selected correctly.}. 
The ground-truth labels are available for the train and validation sets~\cite{R2C}. 
Therefore, we reported the performance of our best model on the testing set once and performed other experiments on the validation set.
\subsection{Implementation Details}
The PyTorch toolkit~\cite{PyTorch} is leveraged to implement our models, and all the experiments were conducted on a single GeForce RTX 2080 Ti GPU. 
The specific details of each model are shown below.

\textbf{Vanilla attention model.}
As for the input features, we employed the pre-trained object-level features extracted by TAB-VCR~\cite{TAB-VCR} as the visual features, and the pre-trained token embeddings from BERT as the textual features. 
All the trainable parameters are initialized with the default PyTorch settings. 
The training batch size is set to 96 and the alignment loss weight $\lambda$ is set to 1.0. 
The parameters are optimized with the Adam~\cite{Adam} optimizer with an initial learning rate $2\times10^{-3}$.

\textbf{VL-Transformers.}
In order to verify the generalizable capability of our attention alignment mechanism on VL-Transformers, we employ several classical VL-Transformer models~\cite{UNITER, VL-BERT, villa} as the backbone.
For a fair comparison, we strictly followed the original implementations. 

\begin{figure*}
    \centering
    \includegraphics[width=0.9\linewidth]{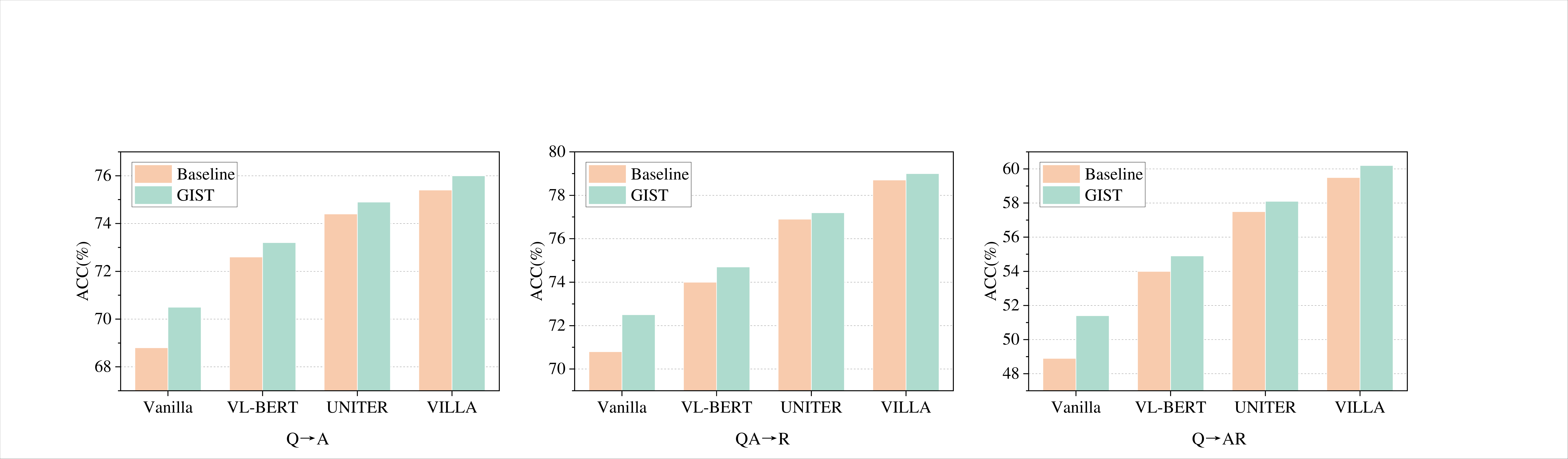}
    \caption{Performance  comparison of different models \emph{w/} and \emph{w/o} our attention alignment module.}
    \label{fig:ablation}
\end{figure*}

\subsection{Overall Performance Comparison}
We evaluated our method by comparing its performance with three kinds of methods: (1) advanced VQA models. (2) Traditional VCR baselines. (3) VL-Transformers in VCR.
The results on both validation and testing sets are reported in Table \ref{tab:overall}, and the key observations are as follows.
\begin{itemize}
    \item First, a significant performance gap exists between traditional strong VQA models and VCR methods, especially for QA$\rightarrow$R. 
    This is because VCR demands higher-order reasoning capability, which differs from the simple recognition in VQA. 
    In addition, predicting the right rationale is even more challenging. 
    \item Second, compared to the task-specific VCR models such as R2C~\cite{R2C} and CCN~\cite{CCN}, VL-Transformers demonstrate advantages in all metrics, indicating the effectiveness of the universal cross-modal representation learned by pre-training.
    \item Lastly, our method achieves the best performance over these state-of-the-art models.
    Especially, for both the vanilla attention and VL-Transformer VCR baselines, with our attention alignment mechanism, they can all achieve improved gains on both validation and test sets. 
    For example, compared with TAB-VCR, an absolute improvement of 1.1$\%$, 1.2$\%$ and 1.6$\%$ can be observed on Q$\rightarrow$A, QA$\rightarrow$R and Q$\rightarrow$AR respectively. 
\end{itemize}

\begin{table}[htb]
    \centering
  \caption{Effectiveness of different alignment losses.}
  \label{tab:align_loss}
  \scalebox{1}{
  \begin{tabular}{llll}
    \toprule
    \makebox[0.25\linewidth][l]{Alignment Loss}    & \makebox[0.17\linewidth][l]{Q$\rightarrow${A}} & \makebox[0.17\linewidth][l]{QA$\rightarrow${R}} & \makebox[0.17\linewidth][l]{Q$\rightarrow${AR}}\\
    \midrule
    Vanilla attention         & 68.8                                  & 70.8                                   & 48.9                         \\
    \emph{\quad w/} Align-Dot  & 70.5 (\textcolor{blue}{+1.7})         & 72.5 (\textcolor{blue}{+1.7})           & 51.4 (\textcolor{blue}{+2.5}) \\
    \emph{\quad w/} Align-Rank & 69.4 (\textcolor{blue}{+0.6})         & 71.9 (\textcolor{blue}{+1.1})           & 50.0 (\textcolor{blue}{+1.1}) \\
    \midrule
    VL-BERT                   & 72.6                                  & 74.0                                   & 54.0                         \\
    \emph{\quad w/} Align-Dot  & 73.2 (\textcolor{blue}{+0.6})         & 74.6 (\textcolor{blue}{+0.6})           & 54.9 (\textcolor{blue}{+0.9}) \\
    \emph{\quad w/} Align-Rank & 73.2 (\textcolor{blue}{+0.6})         & 74.7 (\textcolor{blue}{+0.7})           & 54.9 (\textcolor{blue}{+0.9}) \\
    \midrule
    UNITER                    & 74.4                                  & 76.9                                   & 57.5                         \\
    \emph{\quad w/} Align-Dot  & 74.7 (\textcolor{blue}{+0.3})         & 77.2 (\textcolor{blue}{+0.3})           & 58.0 (\textcolor{blue}{+0.5}) \\
    \emph{\quad w/} Align-Rank & 74.9 (\textcolor{blue}{+0.5})         & 77.0 (\textcolor{blue}{+0.1})           & 58.1 (\textcolor{blue}{+0.6}) \\
    \midrule
    VILLA                     & 75.4                                  & 78.7                                   & 59.5                         \\
    \emph{\quad w/} Align-Dot  & 75.9 (\textcolor{blue}{+0.5})         & 79.0 (\textcolor{blue}{+0.3})           & 60.2 (\textcolor{blue}{+0.7}) \\
    \emph{\quad w/} Align-Rank & 76.0 (\textcolor{blue}{+0.6})         & 78.8 (\textcolor{blue}{+0.1})           & 60.1 (\textcolor{blue}{+0.6}) \\
  \bottomrule
\end{tabular}}
\end{table}

\begin{table}
  \centering
  \caption{Ablation study from our re-attention module on the validation set.}
  \label{tab:re-attention}
  
  \begin{tabular}{lccc}
  \toprule
   {Model}    & {Q$\rightarrow${A}} & {QA$\rightarrow${R}} & {Q$\rightarrow${AR}}\\
  \midrule 
    Full model                 & 70.5                     & 72.5                      & 51.4              \\
    \emph{w/o} token-wise Att  & 70.1                     & 71.5                      & 50.4              \\
    % Query-guided Att           & 67.8                     & 70.3                      & 47.9              \\
    \emph{w/o} Att             & 68.7                     & 71.2                      & 49.1              \\
  \bottomrule
\end{tabular}
\end{table}

\subsection{Ablation Study}
To better illustrate the effectiveness of our model in detail, we conducted experiments on the essential parts of our method and reported the results on the validation set below.

\textbf{Efficacy of the attention alignment.}
Figure~\ref{fig:ablation} demonstrates the effect of our attention alignment mechanism for both the vanilla attention model and VL-Transformers. 
From this figure, we can observe that among all the variants, our attention alignment mechanism consistently improves the base model on all metrics by a large margin. 
Take the vanilla attention model as an example, our attention alignment mechanism boosts it by 1.7\% (Q$\rightarrow$A), 1.7\% (QA$\rightarrow$R), and 2.5\% (Q$\rightarrow$AR).
One main limitation of these baselines is that they neglect the visual consistency and the interactions between the answering and reasoning.
In contrast, our visual attention alignment offers a bridge to connect these two processes and significantly enhances the baseline performance.

Figure~\ref{fig:convergence_analysis} shows the convergence of answering and reasoning accuracy for baselines and our GIST model. 
It can be seen that the performance of baselines increases fairly or faster than our GIST model.
Nevertheless, with more training steps, GIST outperforms all the baselines by a significant margin. 
One possible reason is that the baselines show certain disadvantages in these difficult instances as the model keeps training.
In contrast, when aligning the visual attention between Q$\rightarrow$A and QA$\rightarrow$R, the model learns to leverage more accurate visual information to perform visual understanding, leading to more improvements.

\begin{figure}[t!]
  \centering
  \includegraphics[width=1.0\linewidth]{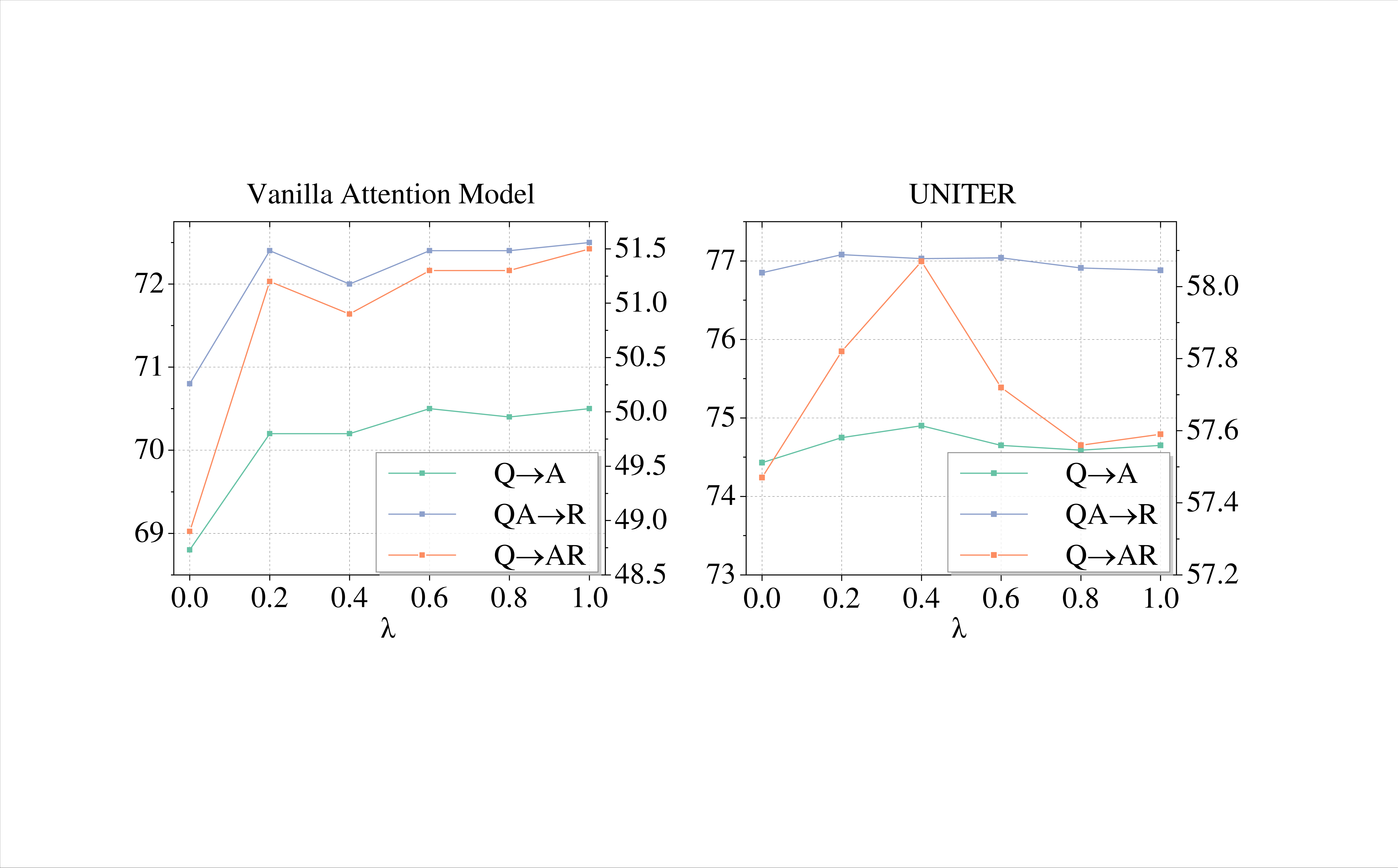}
  \caption{Accuracy performance change with respect to different hyper-parameter $\lambda$.
  Q$\rightarrow${A} and QA$\rightarrow${R} use the left $y$-axis while Q$\rightarrow${AR} uses the right $y$-axis.} \label{fig:lambda}
\end{figure}

\begin{figure}[t!]
  \centering
  \includegraphics[width=1.0\linewidth]{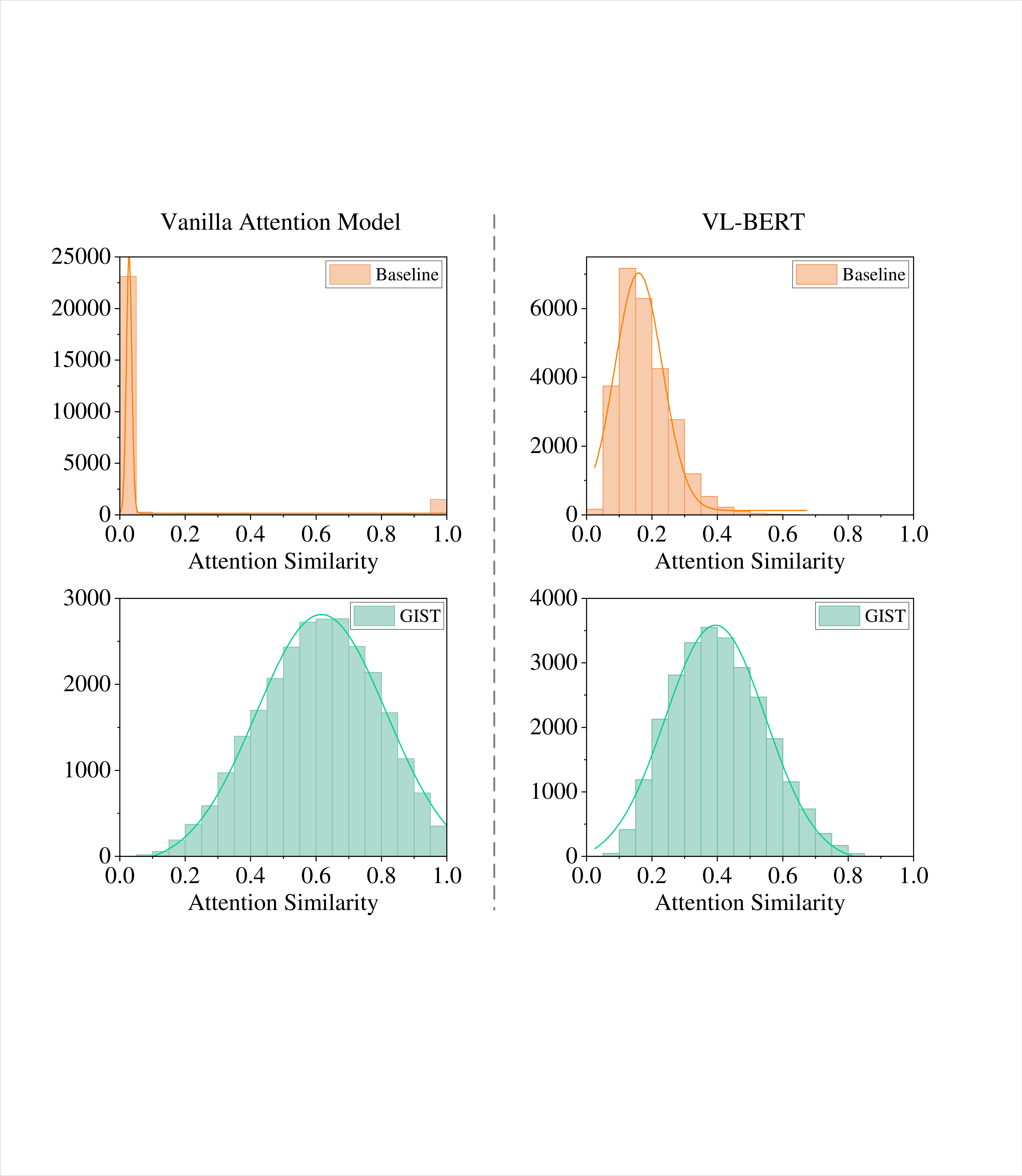}
  \caption{Attention similarity statistics of baselines and our GIST model.} \label{fig:similarity}
\end{figure}

\begin{figure*}[t!]
    \centering
    \includegraphics[width=1\linewidth]{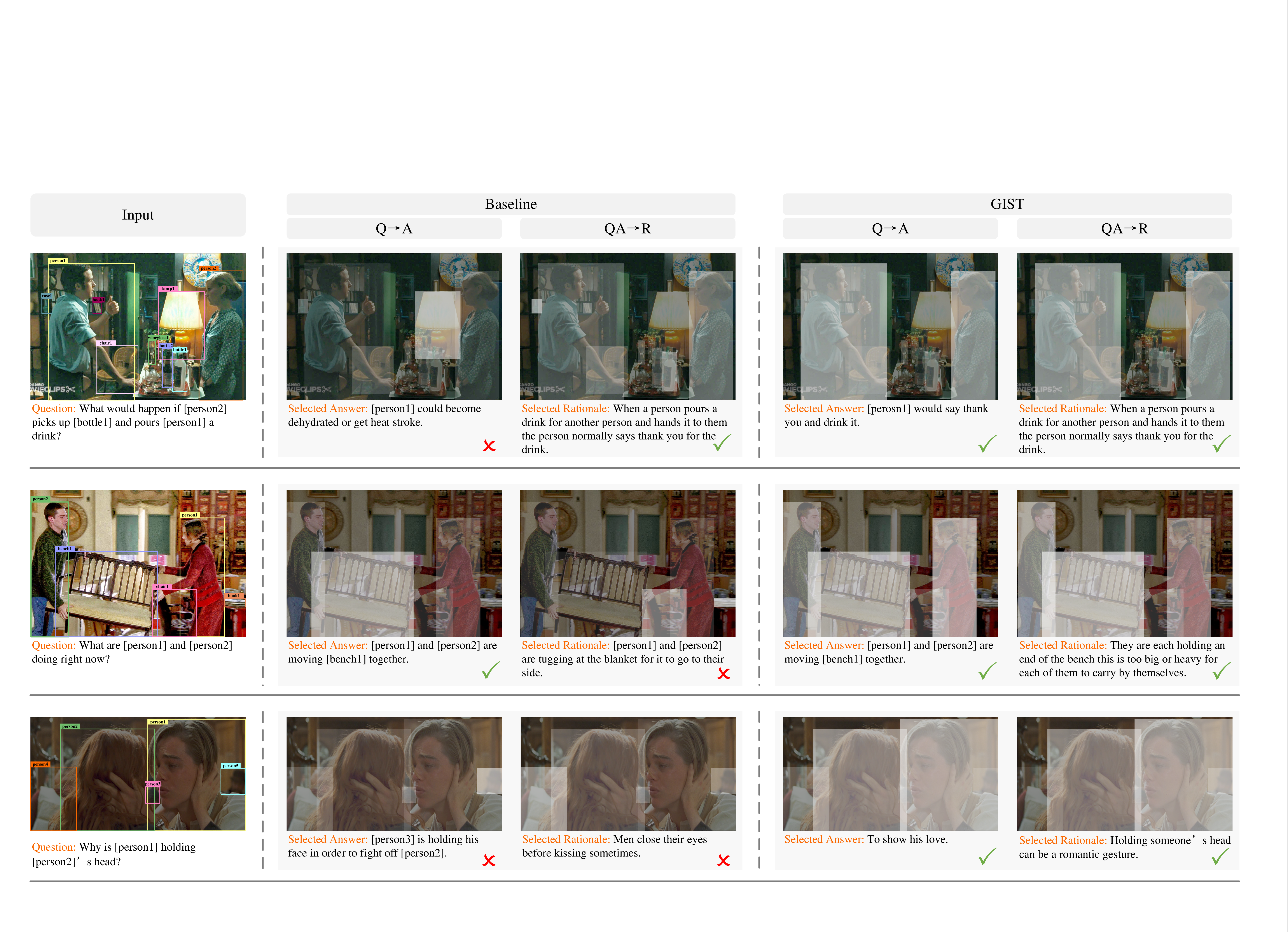}
    \caption{Attention weight distribution of the vanilla attention model and our proposed GIST model.
    The input question and image are shown in the first column, followed by the attention distribution of the baseline in Q$\rightarrow$A and QA$\rightarrow$R (column 2-3), and GIST in Q$\rightarrow$A and QA$\rightarrow$R (column 4-5).}
    \label{fig:case_study}
\end{figure*}

\textbf{Align-Dot v.s. Align-Rank.}
As discussed in Section~\ref{sec:align}, we applied two candidate alignment loss functions to align the two attention maps from the two processes. 
Table \ref{tab:align_loss} reports the results obtained from different alignment losses. 
One can see that both loss functions bring certain performance improvements over respective baselines.
Specifically, for the vanilla attention model, the \emph{Align-Dot}, which is more demanding, can achieve significantly better results. 
We suspect that one possible reason is that the carefully designed re-attention module can capture the attention precisely.

\textbf{Re-Attention of the vanilla attention model.}
In order to investigate the effectiveness of our re-attention module in the vanilla attention model, we designed two variants and reported the results in Table~\ref{tab:re-attention}.
After removing the token-wise attention module from our model, we can observe performance degradation on all three metrics.
It validates that different textual tokens contribute distinctively to visual feature learning.
We then replaced all the attention computation with the mean operation and showed the results in the last row of this table.
One can see that the model performance drops sharply compared with the previous two.

\subsection{Hyper-parameter Study}
In this section, we study the influence of the trade-off hyper-parameter $\lambda$ on the performance of our GIST model. 
The results on both the vanilla attention model and UNITER are shown in Figure~\ref{fig:lambda}. 
As we increase the loss weight $\lambda$, the model performance keeps being enhanced. 
This result demonstrates the effectiveness of our designed attention alignment mechanism. 
However, a too-large weight will lead to a deteriorating result.
For instance, a loss weight larger than 0.4 for UNITER negatively hurts the model performance.

\subsection{Qualitative Results}
It is expected that our attention alignment operation should pull the attention maps from the two processes similarly so that the model prediction can be made according to consistent visual cues.
To justify this, we performed some qualitative results from the following two angles.

\textbf{Similarity of attention maps.}
We leveraged the \emph{Align-Dot} to perform the attention alignment and estimated the attention similarity between Q$\rightarrow$A and QA$\rightarrow$R.
The histogram of the similarity values is shown in Figure~\ref{fig:similarity}.
As can be observed, the attention similarity from baseline models is mostly less than 0.2.
In contrast, our GIST model yields more consistent visual reasoning results as the similarity is increased significantly.

\textbf{Attention map visualization.}
To gain a deeper insight into our attention alignment mechanism, in this section, we also provide some qualitative examples in Figure~\ref{fig:case_study} to compare the attention distribution of the baseline and our method.
Regarding the first instance, the baseline puts more attention on the \emph{lamp} regions while ignoring the critical \emph{person} areas. 
However, though the answer is wrongly predicted, the rationale is unexpectedly selected correctly. 
This supports our argument that the answer prediction and rationale selection should depend on the same evidence otherwise may lead to unpredictable results.
As to the second instance, without the attention alignment, the baseline focuses more on the \emph{chair1} object, which is less relevant to the given question.
Our GIST model corrects this mistake and obtains the right selection for both Q$\rightarrow${A} and QA$\rightarrow${R}.
The last one shows that the attention weights are distributed incorrectly for both processes. 
With the consistent alignment of our method, both Q$\rightarrow${A} and QA$\rightarrow${R} can be accurately predicted.
\section{Conclusion and Future Work}
In this paper, we propose a novel vision attention alignment method to bridge the two intertwined processes in visual commonsense reasoning. 
In particular, a re-attention module is first introduced to model the fine-grained inter-modality interactions, followed by the attention alignment equipped with two alternative loss functions. 
We apply this method to both the conventional vanilla attention model as well as the recent strong VL-Transformers.
Through the qualitative and quantitative experiments on the benchmark dataset, the effectiveness of our proposed method is extensively demonstrated.

In the future, we plan to further explore this view, \ie collaborating the two processes in VCR with a single framework.
More techniques that involve the close connection of these two processes are worth more investigation.

\bibliographystyle{IEEETran}
\bibliography{align}

\begin{IEEEbiography}[{\includegraphics[width=1in,height=1.25in,clip,keepaspectratio]{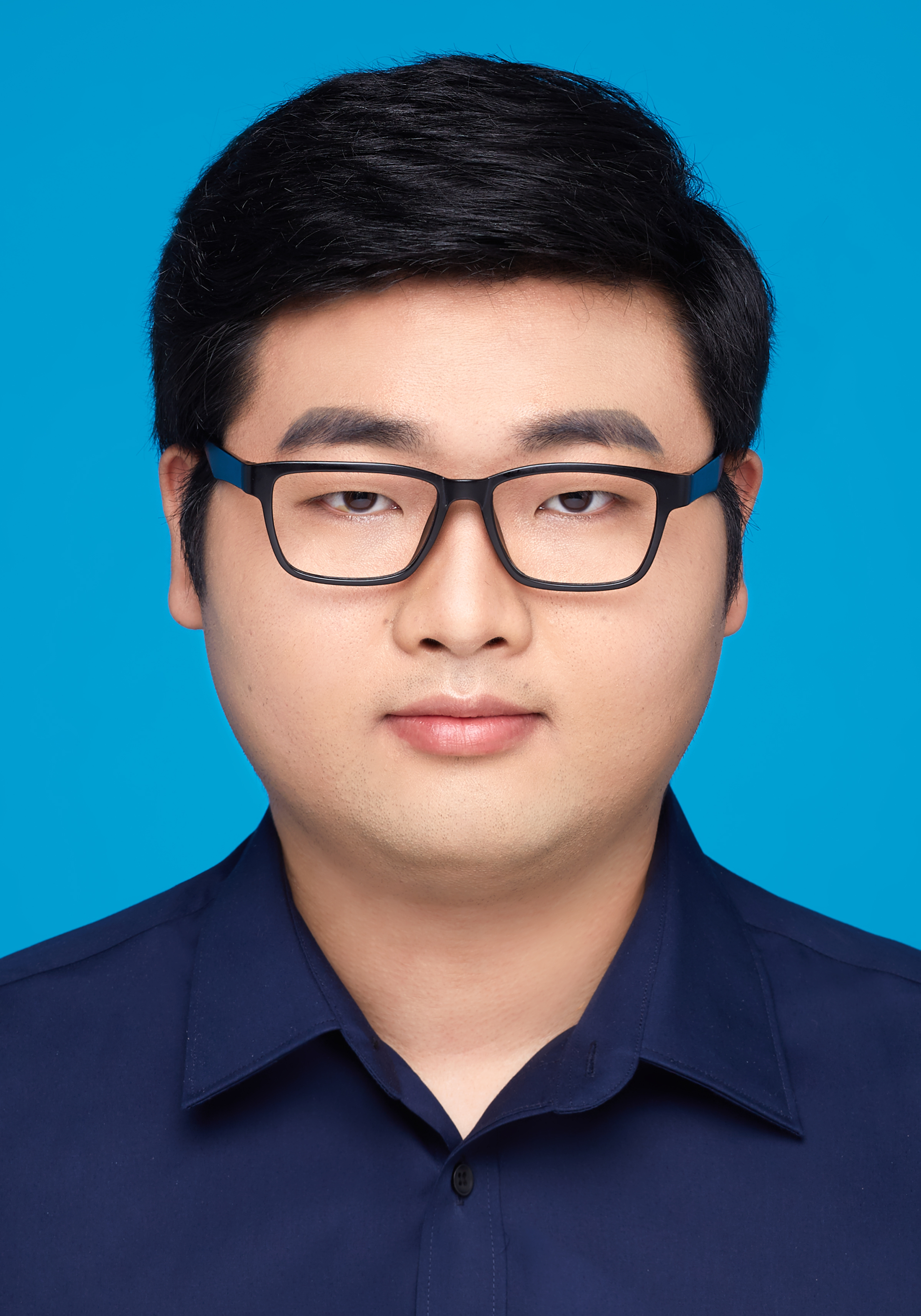}}]{Zhenyang Li}
received the B.Eng. and master degree from Shandong University and University of Chinese Academy of Sciences, respectively. He is currently pursuing the Ph.D. degree with the School of Computer Science and Technology, Shandong University, supervised by Prof. Liqiang Nie. His research interest is multi-modal computing, especially visual question answering.
\end{IEEEbiography}

\begin{IEEEbiography}[{\includegraphics[width=1in,height=1.25in,clip,keepaspectratio]{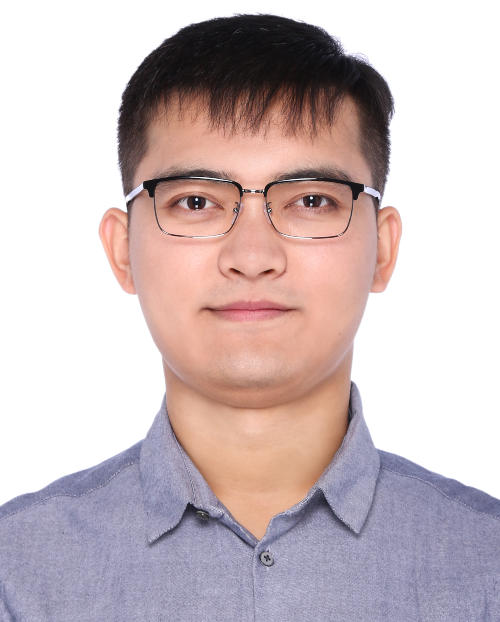}}]{Yangyang Guo}
(Member, IEEE)
is currently a research fellow with the National University of Singapore. 
He has authored or co-authored several papers in top journals, such as IEEE TIP, TMM, TKDE, TNNLS, and ACM TOIS. 
He is a Regular Reviewer for journals, including IEEE TIP, TMM, TKDE, TCSVT; ACM TOIS, and ToMM. 
He was the recipient as an outstanding reviewer for IEEE TMM and WSDM 2022.
\end{IEEEbiography}

\begin{IEEEbiography}[{\includegraphics[width=1in,height=1.25in,clip,keepaspectratio]{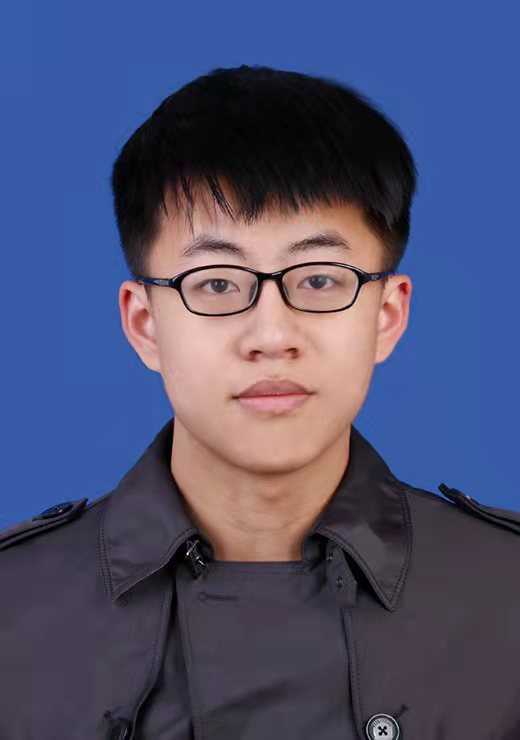}}]{Kejie Wang}
is currently pursuing the B.Eng. degree in computer science from the Shandong University. His research interests include visual question answering and computer vision.
\end{IEEEbiography}

\begin{IEEEbiography}[{\includegraphics[width=1in,height=1.25in,clip,keepaspectratio]{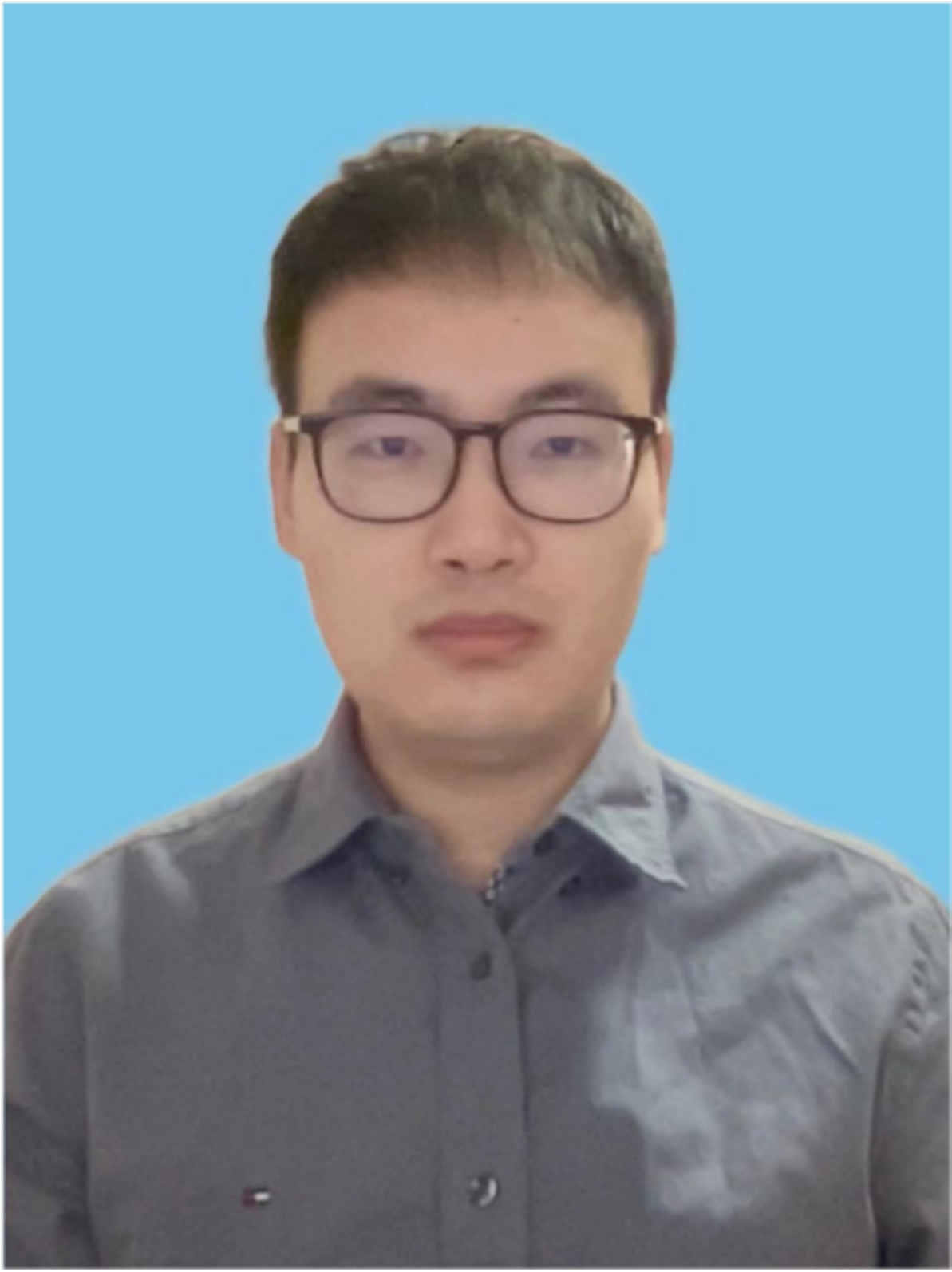}}]{Fan Liu}
(Member, IEEE)
is currently a Research Fellow with the School of Computing, National University of Singapore (NUS). He received the Ph.D degree from Shandong University in China. His research interests lie primarily in multimedia search and recommendation. His work has been published in a set of top forums, including ACM SIGIR, MM, WWW, TKDE, TOIS and TMM. He has served as the pc member for several top conferences, like ACM MM, SIGKDD, WSDM, and the reviewer for journals including TKDE, TMM, IPM, INS.
\end{IEEEbiography}

\begin{IEEEbiography}[{\includegraphics[width=1in,height=1.25in,clip,keepaspectratio]{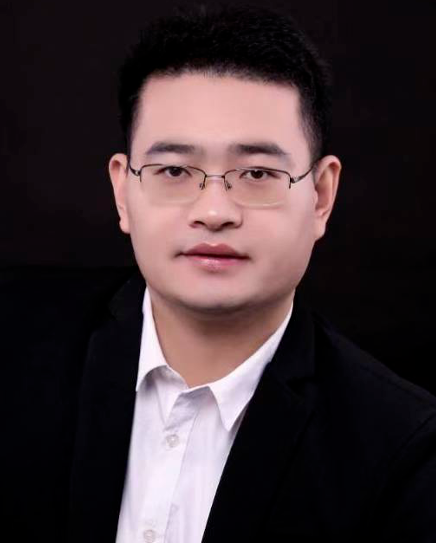}}]{Liqiang Nie}
(Senior Member, IEEE) received the B.Eng. degree from Xi’an Jiaotong University and the Ph.D. degree from the National University of Singapore (NUS). 
He is currently a Professor and the dean of the School of Computer Science and Technology, Harbin Institute of Technology (Shenzhen). 
His research interests lie primarily in multimedia computing and information retrieval.  
He has co-authored more than 200 articles and four books and received more than 14,000 Google Scholar citations. 
He is an AE of IEEE TKDE, IEEE TMM, IEEE TCSVT, ACM ToMM, and Information Science. 
Meanwhile, he is the regular area chair of ACM MM, NeurIPS, IJCAI, and AAAI. 
He is a member of ICME steering committee. 
He has received many awards, like ACM MM and SIGIR best paper honorable mention in 2019, SIGMM rising star in 2020, TR35 China 2020, DAMO Academy Young Fellow in 2020, and SIGIR best student paper in 2021.
\end{IEEEbiography}

\begin{IEEEbiography}[{\includegraphics[width=1in,height=1.25in,clip,keepaspectratio]{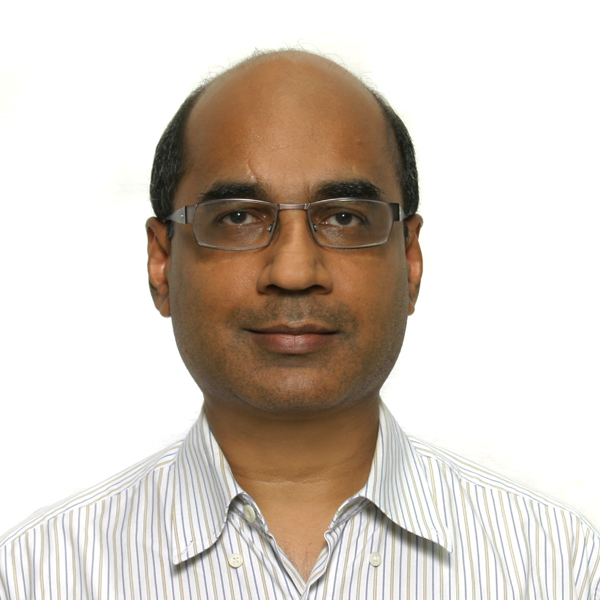}}]{Mohan Kankanhalli}
(Fellow, IEEE)
received the B.Tech. degree from IIT Kharagpur and the M.S. \& Ph.D. degrees from the Rensselaer Polytechnic Institute. 
He is currently the Provost’s Chair Professor at the Department of Computer Science, National University of Singapore. 
He is the Director of N-CRiPT and also the Deputy Executive Chairman of AI Singapore (Singapore's national AI program). 
His current research interests include multimedia computing, multimedia security and privacy, image/video processing, and social media analysis.
.
\end{IEEEbiography}

\end{document}